\pdfoutput=1
\documentclass[11pt]{article}

\usepackage[margin=1.1in]{geometry}
\usepackage[T1]{fontenc}
\usepackage[utf8]{inputenc}
\usepackage{lmodern}
\usepackage{microtype}
\usepackage{booktabs}
\usepackage{longtable}
\usepackage{amsmath}
\usepackage[numbers,sort&compress]{natbib}
\usepackage{xcolor}
\usepackage[colorlinks=true,linkcolor=blue!60!black,citecolor=blue!60!black,urlcolor=blue!60!black]{hyperref}
\usepackage{caption}
\title{\textbf{How Language Models Learn Taste:\\Brand Preference in LLMs}}
\author{Tapan Parikh\\Cornell Tech\\\texttt{tsp53@cornell.edu}}
\date{October 2026}

\begin{document}
\maketitle

\begin{abstract}
\noindent
Asked to name a running shoe brand, language models say Nike. Asked to recommend one, they say Brooks. This paper asks
where models learn that taste. Across 102 models and 42 brand categories, the brand a model picks when recommending
differs from the brand it names in 26. The field moves together to a shared recommendation. The Named brands are market leaders, and the Picked brands
are mostly niche and specialist. Three open training pipelines show where this forms. Pretraining already attaches the recommended brand to advice. Supervised fine-tuning
teaches the model to answer and mostly leaves the pick unchanged. Preference optimization (DPO) teaches the model which answer
is better, and the pick moves in all three pipelines, by 11 to 24 points. The preference data contains almost no
brands. DPO moves the model's judgments toward a second set of associations even when the preferred brand is not in the preference data, and the effect reaches beyond brands.
Asked to choose a sauce, models switch from ketchup to pesto. We do not know exactly how these preferences arise. We
suspect they come from the judge models that decide which answers count as better during preference training. For
brands, a recommendation depends less on what a model knows about them than on whether they fit this taste.
\end{abstract}

\section{Introduction}
\label{sec:intro}

Ask a language model to name a running shoe brand and it says Nike. Ask it to recommend one and it says Brooks. The
first question asks what comes to mind. The second asks for a judgment.

Across 102 models, the brand they pick when recommending differs from the brand they name in 26 of 42 categories. The
models also converge on the same Picked brand. Delta gives way to Singapore Airlines, Chrome to Firefox and Google to
DuckDuckGo. Every major lab makes the move. The Named brands are market leaders, and the Picked brands are mostly
niche and specialist.

This paper asks where models learn this taste. We use open models that release a checkpoint after each stage of
training. Pretraining supplies the candidates. A base model already leans to the recommended brand when a sentence is
phrased as advice. Supervised fine-tuning (SFT) teaches the model to answer but mostly does not change the pick. Direct
preference optimization (DPO) teaches the model which of two answers is better, and the pick moves at DPO in all
three pipelines. Reinforcement learning (RL) leaves it there.

DPO's data holds almost no brand comparisons, so the models do not learn these picks from brand examples. The shift
also appears outside brands. Asked to name a sauce, models say ketchup. Asked to choose one, OLMo 3 7B says pesto 5\%
of the time as a base model and 67\% of the time after DPO. For brands, DPO moves recommendations away from the brand
the base model finds most obvious. We do not know exactly how these associations form. We suspect they come from how
answers are judged during DPO. In our initial probes, an assistant answering these questions did not search the web.

\section{Related work}
\label{sec:related}

\paragraph{Brand recommendations.} Audits of production assistants measure which brands surface and how stable they
are. Paraphrases of one buying intent share few recommendations \citep{jack2026paraphrase}, retrieval moves small brands
more than priors do \citep{tannenbaum2026stage}, and well-known brands win ties on identical specifications
\citep{chu2026brands}. Audits average over wordings \citep{zatuchin2026dice}. Changing the frame of a question changes
model answers \citep{trhlik2026multitudes,rabbani2026fact,tonneau2026cues}.

\paragraph{Convergence and training stages.} Models converge on open-ended outputs \citep{jiang2025hivemind} and on
one-word defaults \citep{parikh2026census}. On the same open pipelines we use, output homogeneity is traced to SFT
\citep{fortier2026convergence} and diversity loss to DPO \citep{karouzos2026where}. On Tulu, preference data matters more
than the optimization algorithm \citep{ivison2024unpacking}.

\paragraph{Transmission between models.} Students inherit traits of the models that generate their data
\citep{shimabucoro2024llmsee,cloud2025subliminal}, and a generic preference dataset can install preferences
\citep{adenali2026subliminal}. Sycophantic agreement appears at DPO on OLMo 3 7B \citep{blank2026sycophantic}. LLM
judges favour students of their own lineage \citep{li2025leakage}. Human raters reward assertive answers
\citep{hosking2023human}.

\section{Setup}
\label{sec:method}

\paragraph{Questions.} The brand battery has 42 categories, from soda to cloud providers. Each category is asked with
three verbs, each asking for one name:
\begin{itemize}
  \item Name: ``Name a running shoe brand. Reply with the name only.''
  \item Choose: ``Choose a running shoe brand. Reply with the name only.''
  \item Recommend: ``Can you recommend a running shoe brand? Reply with the name only.''
\end{itemize}
The two-turn pick asks ``Can you recommend a running shoe brand?'' and then ``Which one would you pick?'' in the same
conversation. We call the answers the Named, Chosen, Recommended and Picked brands. Appendix~\ref{app:panel} checks the
three verbs on open-ended replies without the one-name instruction.

\paragraph{Panel.} We query 102 models through OpenRouter at temperature 1, as each is served. The one-name questions
run 8 times per model and category, the two-turn pick 4 times. Twenty-five hybrid models were also run with reasoning
turned off. Runs were collected between 2026-09-30 and 2026-10-10. Perplexity's Sonar answers from live search and is
left out.

\paragraph{Scoring.} A one-name reply counts if it is one name, with no list and no lead-in. A two-turn reply is
reduced to the brand it commits to. A list, a question back or a refusal counts as no pick, which happens in 17\% of
two-turn replies. On three hand-checked samples the committed-brand rules agreed with a human reading on 87 to 97\% of
replies. Aliases map variant spellings and product names to one brand.

\paragraph{Measures.} For each question we report three measures. Two describe the field and one describes each model.
The first counts the categories where the field's most common brand differs from its Named brand. The second is the
mean share of answers that go to the field's most common brand. The third asks whether each model holds its own
ground: how often its answer is the brand it gives itself when asked to name one, among replies that name a brand.
Appendix~\ref{app:panel} reads each contrast against a split-half floor.

\section{The field names one brand and picks another}
\label{sec:panel}

Table~\ref{tab:ladder} goes from naming to picking. Under Name every category has a clear consensus, and a model gives
its own Named brand 89\% of the time. Each step toward a judgment changes the field's brand in more categories: Choose
in 6 of 42, Recommend in 16, and the two-turn pick in 26. Over the same steps, a model's own Named brand falls from 89\%
of its answers to 31\% of its picks.

\begin{table}[t]
\centering\small
\caption{From naming to picking (102 models, 42 categories). Consensus moved: categories where the most common
answer across models differs from the most common answer under Name. Answer unchanged: share of a model's answers
that match its own most common answer under Name (under Name itself, how consistently it repeats itself).}
\label{tab:ladder}
\begin{tabular}{lrr}
\toprule
Question & Consensus moved & Answer unchanged \\
\midrule

Name & 0 of 42 & 89\% \\
Choose & 6 of 42 & 73\% \\
Recommend & 16 of 42 & 57\% \\
Recommend, then pick & 26 of 42 & 31\% \\

\bottomrule
\end{tabular}
\end{table}

Each question has its own consensus. The most common brand takes 76\% of answers under Name and between 60 and 69\%
for the other questions. The models move together to a new brand. Table~\ref{tab:flips} lists the 26 categories where the Picked
brand differs from the Named brand. Nike becomes Brooks, Chrome becomes Firefox, Delta becomes Singapore Airlines and
WhatsApp becomes Signal.

The Named brands are market leaders: Google, Walmart, McDonald's, Facebook, Verizon. The Picked brands are mostly
smaller brands with a specialist or insider following: DuckDuckGo, Aldi, Sierra Nevada, Reddit, Signal, Brooks. A few
are premium brands, such as Hermès over Chanel and Hyatt over Marriott.

\begin{table}[t]
\centering\small
\caption{The field's brand for each question, in the 26 categories where the Picked brand differs from the Named
brand.}
\label{tab:flips}
\setlength{\tabcolsep}{4pt}
\begin{tabular}{lllll}
\toprule
Category & Named & Chosen & Recommended & Picked \\
\midrule

AI assistant & Siri & Claude & ChatGPT & Claude \\
airline & Delta & Delta & Singapore Airlines & Singapore Airlines \\
bank & Chase & Chase & Chase & Ally Bank \\
beer & Heineken & Guinness & Guinness & Sierra Nevada \\
browser & Chrome & Firefox & Firefox & Firefox \\
camera & Canon & Canon & Canon & Sony \\
coffee brand & Starbucks & Starbucks & Lavazza & Stumptown \\
coffee chain & Starbucks & Starbucks & Starbucks & Peet's Coffee \\
credit card & Visa & Visa & Chase & Chase \\
electric car & Tesla & Tesla & Tesla & Hyundai \\
fashion & Nike & Nike & Patagonia & Uniqlo \\
fast food & McDonald's & McDonald's & McDonald's & Chick-fil-A \\
hotel & Marriott & Marriott & Marriott & Hyatt \\
laptop & Dell & Dell & Lenovo & Apple \\
luxury & Chanel & Hermès & Hermès & Hermès \\
messaging app & WhatsApp & WhatsApp & Signal & Signal \\
mobile carrier & Verizon & Verizon & T-Mobile & T-Mobile \\
project management software & Trello & Asana & Asana & Asana \\
running shoe & Nike & Nike & Brooks & Brooks \\
search engine & Google & Google & Google & DuckDuckGo \\
smartphone & Samsung & Samsung & Samsung & Google Pixel \\
snack brand & Doritos & Doritos & KIND & KIND \\
sneaker & Nike & Nike & Nike & New Balance \\
social network & Facebook & Instagram & Mastodon & Reddit \\
supermarket & Walmart & Walmart & Walmart & Aldi \\
watch & Rolex & Rolex & Seiko & Seiko \\

\bottomrule
\end{tabular}
\end{table}

In the two-turn pick, a model's own Named brand appears in its recommendation list 83\% of the time and is its pick
31\% of the time. When the Named brand is listed, it comes first in 60\% of lists and is picked in 37\%. A random
draw from the list, which has a median of four brands, would pick it 30\% of the time. Models do not reject their
default. They demote it from first place, and the field's other picks converge on the same brand.

In each of the 12 labs with at least three models, the two-turn pick changes the lab's consensus brand in 55 to 74\% of categories. Two halves of the Name
runs disagree on the consensus brand in no category (Appendix~\ref{app:panel}).

Labs favour their own AI assistant. Asked to recommend an AI assistant, Anthropic's models name Claude in 92\% of their
answers, and other labs' models name it in 15\% of theirs. This inflates the AI assistant row of
Table~\ref{tab:flips}. Labs do not favour their other products. Google's models recommend Google search and Chrome
less often than other labs' models do.

\paragraph{An assistant with search answered from memory.} An assistant that searches the web could answer differently. In a spot
check, Claude Opus 5.5 answered the two-turn question in all 42 categories with web search available and searched in
none of them. It left the Named brand in 25 of the 26 categories where the panel does, and kept the panel's brand in
14 of the 16 where the panel holds.

\section{Taste forms when a model learns which answer is better}
\label{sec:stages}

Three pipelines release a checkpoint after each stage of training: OLMo 3 7B, OLMo 3.1 32B and Tulu 3 8B. Each has a
base model, an SFT stage, a DPO stage and an RL stage. Nemotron 3.5 Lightning releases a base and a final model. We read
the base stage as text completion and the tuned stages through their chat format with no system prompt. We score by
log-probability, which measures every stage on every category without sampling. A brand's score is the summed
log-probability of its name as the continuation, combined over its common spellings. Comparisons are between two
brands at the same stage, or of one brand across stages, so a long name such as Singapore Airlines scores against
itself. Sampled picks and log-probabilities mostly agree. The exception is OLMo 3 7B, whose sampled picks move at RL
rather than DPO (Appendix~\ref{app:stages}).

\paragraph{Pretraining supplies the recommended brand.} A base model cannot be asked for advice, but it can complete an
advice sentence. Given ``If you want a good web browser, I recommend'', Tulu's base model continues with Firefox in 13
of 20 samples. Given ``A well-known web browser is'', it continues with Chrome in 10. Over the 16 categories where the
panel's Recommended brand differs from its Named brand, advice phrasing raises the Recommended brand's share in 7 to 10
categories, depending on the base model, and lowers it in at most one. Log-probabilities agree in every base model
(Appendix~\ref{app:stages}).

\paragraph{Learning to answer doesn't change the answer.} To compare stages on the same choice, we give every stage
the same three-item list and ask ``Which one would you pick?''. The list holds the Named brand, the Picked brand and a
third brand the panel recommends. The reply is started with ``I'd pick'', and we read the probability of each brand.
Each list is shown in both orders, because base models favour the brand listed first. Table~\ref{tab:stages} gives the
share for the Picked brand over the Named brand. After SFT it sits near the base value in all three pipelines. On
direct answers OLMo 3.1 32B already moves at SFT (Appendix~\ref{app:stages}).

\begin{table}[t]
\centering\small
\caption{Fixed-list pick by stage: probability of picking the panel's Picked brand over its Named brand, averaged over
both list orders and over the 26 categories where the two differ. Last column: the SFT-to-DPO change with a 90\%
bootstrap interval over categories.}
\label{tab:stages}
\begin{tabular}{lrrrrl}
\toprule
Pipeline & Base & SFT & DPO & RL & SFT $\to$ DPO \\
\midrule

OLMo 3 7B & 58\% & 52\% & 76\% & 74\% & +24 [+16, +33] \\
Tulu 3 8B & 54\% & 58\% & 69\% & 68\% & +11 [+6, +16] \\
OLMo 3.1 32B & 63\% & 57\% & 73\% & 73\% & +16 [+7, +25] \\

\bottomrule
\end{tabular}
\end{table}

\paragraph{Learning the better answer moves it.} DPO raises the Picked brand's share in all three pipelines, by 24
points in OLMo 3 7B, 11 in Tulu 3 8B and 16 in OLMo 3.1 32B. RL leaves it where DPO put it. Nemotron, which has only a
base and a final model, rises from 58\% to 79\%.

The move does not depend on which brands the panel picks. Over all 42 categories, we compare each brand's
log-probability under Name at the base stage with how much DPO raises it under Recommend. The correlation is negative
in all three pipelines ($-0.16$, $-0.14$ and $-0.23$, each interval below zero). DPO raises the brands the base model
found less obvious. DPO does not lower obvious answers everywhere. In the 16 categories where the panel keeps the
Named brand, its share under Recommend does not fall at DPO in any pipeline. In the 26 where the panel moves, it falls
in all three, significantly in Tulu 3 8B and OLMo 3.1 32B.

\paragraph{The same step on the models' own lists.} We repeat the measure on the SFT stage's own replies to ``Can you
recommend\ldots'', keeping those that name both brands (89 to 146 lists per pipeline). The DPO step remains, at $+14$,
$+9$ and $+14$ points. Appendix~\ref{app:stages} compares log-probabilities with sampled picks and checks a known side
effect of DPO.

\section{What preference optimization teaches}
\label{sec:general}

\paragraph{The data holds almost no brands.} OLMo 3's SFT and DPO data and Tulu 3's SFT and preference mixtures are
public. We searched them for requests that name one of the 26 categories alongside an advice word such as
``recommend'', ``best'' or ``which'', and read the first brand each answer names. In 260,000 OLMo 3 DPO pairs, the
chosen answer names the Picked brand 3 times and the rejected answer names it once. Tulu 3 is as sparse
(Appendix~\ref{app:stages}). The search finds almost no explicit examples from which DPO could learn these picks.
It does not rule out indirect associations elsewhere in the data.

\paragraph{The same step moves judgments outside brands.} The census asks everyday categories, such as a tree, a smell
or an animal, with Name and Choose \citep{parikh2026census}. In 23 of them the panel's Choose answer differs from its
Name answer. Asked for a sauce, the panel names ketchup and chooses pesto. We read the same log-probability shift on
these 23 categories (Table~\ref{tab:census}). Base models barely separate the two verbs. DPO moves Choose toward the
panel's answer in all three pipelines. For OLMo 3 7B asked to choose a sauce, pesto goes from 5\% at base to 27\% after
SFT and 67\% after DPO, and ketchup falls from 72\% to 6\%. OLMo 3.1 32B is the exception at SFT. Its shift already
rises there, on these categories and on brands, and DPO adds to it.

\begin{table}[t]
\centering\small
\caption{Census Choose: how far ``Choose'' moves the odds toward the panel's Choose answer, relative to ``Name'' (nats),
over the 23 categories where the two differ. Last column: the SFT-to-DPO change, 90\% interval.}
\label{tab:census}
\begin{tabular}{lrrrrl}
\toprule
Pipeline & Base & SFT & DPO & RL & SFT $\to$ DPO \\
\midrule

OLMo 3 7B & -0.00 & +0.12 & +0.50 & +0.44 & +0.39 [+0.11, +0.67] \\
Tulu 3 8B & +0.08 & +0.17 & +0.28 & +0.15 & +0.11 [+0.00, +0.21] \\
OLMo 3.1 32B & +0.30 & +1.30 & +1.47 & +1.47 & +0.17 [+0.07, +0.27] \\

\bottomrule
\end{tabular}
\end{table}

\paragraph{Other models define the better answer.} Half of OLMo 3's DPO pairs are judged by an LLM. The other half pair
a response from Qwen 3 32B, as chosen, with one from a much smaller model, as rejected \citep{geng2025delta}. Tulu 3's
pairs are rated by GPT-4o \citep{lambert2024tulu}. In both pipelines other models decide which answer is better, and
the student applies it to situations mostly unrelated to the preference data.

\paragraph{Teacher's favourite.} Tulu 3's preference data has one judge. Answers from a pool of 22 models are rated by
GPT-4o \citep{lambert2024tulu}, and the top-rated answer becomes the chosen one. Where GPT-4o recommends a different
brand from Tulu's own base-model favourite, we give each stage a list with both, in both orders. DPO moves the pick
toward GPT-4o's brand, by 8 points (90\% interval 2 to 14). The GPT-4o we query is the current release, not the
2024-08-06 snapshot that labelled Tulu's data, so its brands approximate the judge's. OLMo 3's data has no single teacher, so the test applies
to Tulu only (Appendix~\ref{app:stages}).

\section{Conclusion}
\label{sec:discussion}

Language models name the market leader but recommend a different brand, and different models agree on which one. On
open pipelines the recommended brand is present after pretraining, attached to advice. SFT mostly leaves the pick
unchanged.
DPO moves the model to the recommended brand, using preference data that contains almost no brands.

DPO moves the model toward a second set of associations that we can only partly describe. The Picked brands include
specialist and enthusiast brands, such as Brooks, Sierra Nevada and Stumptown, independent or privacy-minded ones,
such as DuckDuckGo, Signal and Firefox, and value or premium brands chosen for quality, such as Aldi, Uniqlo, Hermès
and Seiko. Underneath these switches is an implicit notion of taste that we can only speculate about for now. In Tulu,
the pick moves toward the brand its judge model would recommend. That is a partial answer, because the preference data
contains almost no examples of these comparisons.

For a brand, this makes a recommendation hard to earn directly. It depends less on what a model knows about the brand
and more on whether the brand fits this notion of taste. In marketing terms this is a question of brand associations
and positioning \citep{keller1993equity,aaker1997personality}. How a brand can create
these associations for itself is a question for future work.

\section{Limitations}
\label{sec:limits}

Our questions are short and carry no personal context. Measures on short prompts can fail to predict realistic use
\citep{lum2025trick}. Audits of production assistants measure real requests and complement this work.

The three staged pipelines come from one lab and share much of their recipe. The panel shows the same switch across
102 served models from every major lab, so the outcome holds beyond these pipelines even though the route is shown
only on them. Nemotron agrees from base to final but has no intermediate stages. The 32B and Nemotron run at 8-bit
precision. For OLMo 3 7B, sampled picks place the change at RL (Appendix~\ref{app:stages}). The panel is a dated
specimen of models as served between 2026-09-30 and 2026-10-10.

\section*{Note on AI usage}
This work was done with Claude (Opus 5.5), which helped run the probes, build the analysis and
draft the text; the research questions and interpretation are the author's. Several Claude
models are also subjects of the study.

\bibliographystyle{plainnat}
\bibliography{references}

\clearpage
\appendix

\section{Panel checks}
\label{app:panel}

\paragraph{Floors and the one-name instruction.} Table~\ref{tab:factors} reads each verb contrast against a split-half floor. Two
halves of the Name runs disagree on the consensus brand in no category, while Name and Recommend disagree in 38\% of
categories. Dropping the one-name instruction leaves Name nearly unchanged and moves Choose and Recommend in about a third of
categories. The verb effect holds in both forms.

\begin{table}[h]
\centering\small
\caption{Verb and one-name effects against split-half floors. ``Moved'' is the share of categories whose consensus brand
changes. ``Kept'' is how often a model's modal answer survives the change.}
\label{tab:factors}
\begin{tabular}{lrrrr}
\toprule
Contrast & Moved & Floor & Kept & Floor \\
\midrule

Name $\to$ Choose & 14\% & 0\% & 73\% & 85\% \\
Name $\to$ Recommend & 38\% & 0\% & 57\% & 85\% \\
free Name $\to$ two-turn pick & 64\% & 1\% & 30\% & 84\% \\
one name $\to$ free Name & 7\% & 0\% & 80\% & 85\% \\
one name $\to$ free Choose & 29\% & 0\% & 61\% & 82\% \\
one name $\to$ free Recommend & 31\% & 0\% & 60\% & 82\% \\

\bottomrule
\end{tabular}
\end{table}

\paragraph{Paraphrase.} Rewording Recommend twice, on a 14-model subset, leaves its consensus brand unchanged in 34 to
36 of 39 categories.

\paragraph{One-turn pick.} The one-turn question ``Which running shoe brand would you pick?'' moves the consensus in
15 of 42 categories with the one-name instruction. Without it, it names no brand in 58\% of replies, so it measures refusal more than
preference.

\section{Stage checks}
\label{app:stages}

\paragraph{The direct answer.} Table~\ref{tab:shift} reads how far ``Can you recommend\ldots'' moves the odds from the
Named brand toward the Picked brand, relative to ``Name\ldots''. The shift is already present in every base model. It
rises at DPO in all three pipelines, and two rewordings of Recommend rise with it. OLMo 3.1 32B also shows a large
rise at SFT in the direct answer, on brands and on census categories alike. Its pick from a list does not rise at SFT.

\begin{table}[h]
\centering\small
\caption{How far ``Can you recommend\ldots'' moves the odds from the Named brand toward the Picked brand, relative to
``Name\ldots'' (nats), by stage. Last column: the SFT-to-DPO change, 90\% interval.}
\label{tab:shift}
\begin{tabular}{lrrrrl}
\toprule
Pipeline & Base & SFT & DPO & RL & SFT $\to$ DPO \\
\midrule

OLMo 3 7B & +1.48 & +1.08 & +1.44 & +1.47 & +0.36 [+0.08, +0.62] \\
Tulu 3 8B & +1.76 & +1.71 & +2.47 & +2.25 & +0.76 [+0.35, +1.20] \\
OLMo 3.1 32B & +0.79 & +3.97 & +4.88 & +4.66 & +0.92 [+0.58, +1.28] \\

\bottomrule
\end{tabular}
\end{table}

\paragraph{Sampled picks.} Sampling picks from the SFT stage's own lists agrees with the log-probabilities within a
few points everywhere except OLMo 3 7B at DPO (Table~\ref{tab:agree}). There, sampled picks do not move until RL.
First-token probabilities and generated answers are known to diverge in tuned models \citep{wang2024answerc}.

\begin{table}[h]
\centering\small
\caption{The pick on the SFT stage's own lists: sampled picks (5 per list) against log-probabilities.}
\label{tab:agree}
\begin{tabular}{llrr}
\toprule
Pipeline & Stage & Sampled & Log-probability \\
\midrule

OLMo 3 7B & SFT & 34\% & 37\% \\
OLMo 3 7B & DPO & 36\% & 51\% \\
OLMo 3 7B & RL & 42\% & 50\% \\
Tulu 3 8B & SFT & 38\% & 42\% \\
Tulu 3 8B & DPO & 49\% & 51\% \\
Tulu 3 8B & RL & 55\% & 51\% \\
OLMo 3.1 32B & SFT & 46\% & 45\% \\
OLMo 3.1 32B & DPO & 61\% & 58\% \\
OLMo 3.1 32B & RL & 61\% & 58\% \\

\bottomrule
\end{tabular}
\end{table}

\paragraph{The base runner-up.} DPO can concentrate probability on outputs that were already likely
\citep{ren2024dynamics} or move it to unrelated ones \citep{razin2024likelihood}. The first would move a model toward
its base runner-up. The base model's second choice under Name is the panel's Picked brand in only 2 or 3 of 26
categories, and DPO does not favour the runner-up over the pick.

\paragraph{Brands in the post-training data.} Table~\ref{tab:data} counts the first brand named in answers to
recommendation requests in the 26 categories, in OLMo 3's and Tulu 3's post-training data.

\begin{table}[h]
\centering\small
\caption{Brand recommendations in the post-training data, over the 26 categories: first brand named in SFT answers, and
in the chosen and rejected answers of preference pairs.}
\label{tab:data}
\begin{tabular}{lrrrrrr}
\toprule
 & \multicolumn{2}{c}{SFT} & \multicolumn{2}{c}{DPO chosen} & \multicolumn{2}{c}{DPO rejected} \\
Data & Named & Picked & Named & Picked & Named & Picked \\
\midrule

OLMo 3 (Dolci) & 43 & 8 & 5 & 3 & 5 & 1 \\
Tulu 3 & 16 & 2 & 10 & 3 & 11 & 2 \\

\bottomrule
\end{tabular}
\end{table}

\paragraph{The teacher test on OLMo.} Half of OLMo 3's preference pairs set an answer from Qwen 3 32B against one from
a much smaller model, with no judge, and the other half are judged across a dozen models. Taking Qwen 3 32B's brand as
the teacher's, the OLMo 3 7B and OLMo 3.1 32B pipelines move $-4$ and $+3$ points at DPO, and both intervals span zero.

\end{document}